\documentclass[11pt,onecolumn,preprintnumbers,amsmath,amssymb]{revtex4}

\usepackage{graphicx}
\usepackage{dcolumn}
\usepackage{bm}
\usepackage{float}
\usepackage[utf8x]{inputenc}

\begin{document}


\title{A Neural Network Study of Blasius Equation}

\author{Halil Mutuk}
 \email{halilmutuk@gmail.com}
 \affiliation{Physics Department, Faculty of Arts and Sciences, Ondokuz Mayis University, Samsun, Turkey}

\begin{abstract}
In this work we applied a feed forward neural network to solve Blasius equation which is a third-order nonlinear differential equation. Blasius equation is a kind of boundary layer flow. We solved Blasius equation without reducing it into a system of first order equation. Numerical results are presented and a comparison according to some studies is made in the form of their results. Obtained results are found to be in good agreement with the given studies.
\end{abstract}

\maketitle

\section{\label{sec:level1} Introduction}
The study of ordinary differential equations in unbounded domains is an important subject. Many physics and engineering problems can be modelled effectively by unbounded or semi-bounded domains. An unbounded domain include the entire set of real numbers which could present considerable theoretical and practical challenges which are not present in bounded domains. One of the well-known equations arising in fluid mechanics and boundary layer approach is Blasius differential equation. In fluid mechanics, the Blasius equation arises to describe the boundary layer with laminar flow of a fluid over a flat plate with fluid moving by it. Due to being a nonlinear boundary value problem, it cannot be solved analytically. A closed-form solution of Blasius equation is evading. Many methods, techniques or approaches have been used to obtain analytical and numerical solutions for Blasius equation. 

The Blasius equation emerged as a solution of convection equation for a flat plate. Blasius transformed continuity and momentum equations for steady, incompressible laminar flow of a fluid by introducing similarity variable. Two dimensional steady state laminar viscous flow over a semi-infinite flat plate is modelled by the nonlinear two-point boundary value
\begin{eqnarray}
f'''(\eta)+\frac{1}{2}f(\eta)f''(\eta)=0, ~ \eta \geq 0 \\
f(0)=0, ~ f'(0)=0, ~ f'(\infty)=1 \nonumber
\end{eqnarray}
where $\eta$ and $f(\eta)$ are the dimensionless coordinate and the dimensionless stream function, respectively. Blasius gave a solution by using a power series expansion for this problem as  Ref. \cite{1}
\begin{equation}
f(\eta)=\sum_{k=0}^{\infty} (-\frac{1}{2})^k \frac{A_k \sigma^{k+1}}{(3k+2)!} \eta^{3k+2},
\end{equation}
where $A_0=A_1=1$, $\sigma=f''(0)$ and 
\begin{equation}
A_k=\sum_{r=0}^{k-1} \left( \begin{array}{c}
3k-1 \\ 3r
\end{array}                     \right) A_r A_{k-r-1}, ~ k\geq2.
\end{equation}
In Ref. \cite{2}, boundary layer equations including Blasius were studied by some approximation methods. 

Many methods or techniques have been used to obtain the analytical and numerical solutions for this equation. The first numerical solution for Blasius equation was obtained by Howarth in \cite{3} by using Runge-Kutta method. Liao studied Blasius equation by the non-linear approximate technique called Homotopy Analysis Method \cite{4}. Yu and Chen applied differential transformation method to solve Blasius equation \cite{5}.  He used a simple perturbation approach to solve Blasius equation \cite{6}. A Runge-Kutta algorithm was used for numerical solutions \cite{7}. Arikoglu and Ozkol studied inner-outer matching solution by differential transformation method \cite{8}. Wazwaz used variational iteration method for solving two forms of Blasius equation \cite{9}. Tajvidi et al. used Legendre tau method to solve Blasius problem approximately \cite{10}. In Ref. \cite{11}, an approximate analytic solution of the Blasius equation by modified the 4/3 Pade approximant is obtained. Parand et al. studied Sinc-collocation method \cite{12} and Girgin used generalized iterative differential quadrature method for solving Blasius equation \cite{13}. In Ref. \cite{14}, the authors used adomian methods to get the analytical solution of Blasius equation. A comparison among the homotopy perturbation method and the decomposition method with the variational iteration method for solving Blasius equation have been studied in \cite{15}. Xu and Guo applied the fixed point method to obtain the semi-analytical solution to Blasius equation \cite{16}. A predictor corrector two-point block method is proposed to solve the well- know Blasius and Sakiadis flow numerically \cite{17}. All these methods have their own shortcomings.  

The usage of artificial neural networks (ANNs) to solve differential equations has become an interesting topic since two decades. Neural networks (NNs) can solve ordinary and partial differential equations that relies on the function approximation capabilities of feed forward neural networks  and results in the construction of a solution written in a diferentiable, closed analytic form \cite{18}. The motivation behind is that neural networks are function approximators due to Cybenko theorem \cite{19}. This form employs a feed forward neural network as the basic approximation element, whose parameters (weights and biases) are adjusted to minimize an appropriate error function. Training of the network can be done by optimizing techniques which require the computation of the gradient of the error in terms of the network parameters. In this model, a trial solution of the differential equation is written as a sum of two parts: the first part satisfies the  given initial/boundary conditions and contains no adjustable parameters and the second part involves a feed forward neural network to be trained so as to satisfy the differential equation. As a consequence, by construction the trial solution, the initial or boundary conditions are satisfied and the network is trained to satisfy the differential equation \cite{18}. 

A neural network study for differential equation provides some advantages over the existing numerical methods \citep{18}:

\begin{itemize}
\item[$\ast$] The neural network based solution of a differential equation is differentiable
and is in closed analytic form that can be used in any subsequent calculation. On the other hand most other techniques offer a discrete solution or a solution of limited differentiability.

\item[$\ast$] Trial solutions of ANN include a single independent variable regardless of the dimension of the problem.

\item[$\ast$] The neural network based method to solve a differential equation provides a
solution with very good generalization properties.

\item[$\ast$] The solutions are continuous over all the domain of the integration. Traditional numerical methods provide solutions over discrete points and the solution among these points must be interpolated.

\item[$\ast$] The required number of model parameters is far less than any other solution technique and
therefore, compact solution models are obtained, with very low demand on memory space.

\item[$\ast$] The method is general and can be applied to ordinary differential equations (ODEs), systems of ODEs and to partial differential equations (PDEs) as well.

\item[$\ast$] The method is general and can be applied to the systems defined on either orthogonal box boundaries or on irregular arbitrary shaped boundaries.

\item[$\ast$] The method can be realized in hardware, using neuroprocessors, and hence offer the opportunity to tackle in real time difficult differential equation problems arising in many engineering
applications.

\item[$\ast$] The method can also be  implemented on parallel architectures.
\end{itemize}

There are different neural network methods for the solutions of differential equations. These are feed forward neural networks \cite{20} in which the information moves in only one direction forward from the input nodes to output nodes, recurrent neural networks \citep{22} in which information can go back from output nodes to input nodes, radial basis function networks \cite{23} in which three layers (input layer, basis function layer as hidden layer, output layer) exist where each node in the hidden layer represents a Gaussian basis function for all nodes and output node have a linear activation function, Hopfield network \citep{24,25} in which  a set of neurons with unit delay is fully connected to each other and forming a feedback neural network, cellular network \citep{26,27} which features a multidimensional array of neurons and local interconnections among cells, finite element  neural network \cite{28} have the finite element model converted into the parallel network form and  wavelet element network \citep{29} which is an alternative to feed forward neural network for approximating arbitrary nonlinear functions as an alternative and have waveleons instead of neurons.  Among these neural network methods, feed forward neural networks are popular tools due to their structural flexibility, good representation capabilities and availability of a large number of training algorithms \cite{20}.

In this work we studied Blasius equation  via feed forward neural networks.  In section \ref{sec:level2} we introduce our formalism of neural network frameworks. In section \ref{sec:level3} we give our results and in section \ref{sec:level4} we summarize our results.

\section{\label{sec:level2}Formalism of Feed Forward Neural Networks}

Artificial neural networks are information processing structures that are parallel distributed in the form of a directed graph. A directed graph is composed of set of points called nodes and directed line  segments called links. ANNs are inspired by the way of process information of brain or in more general terms, by the way of biological nervous systems. 

\begin{figure}[H]
\centering
\includegraphics[width=5.8in]{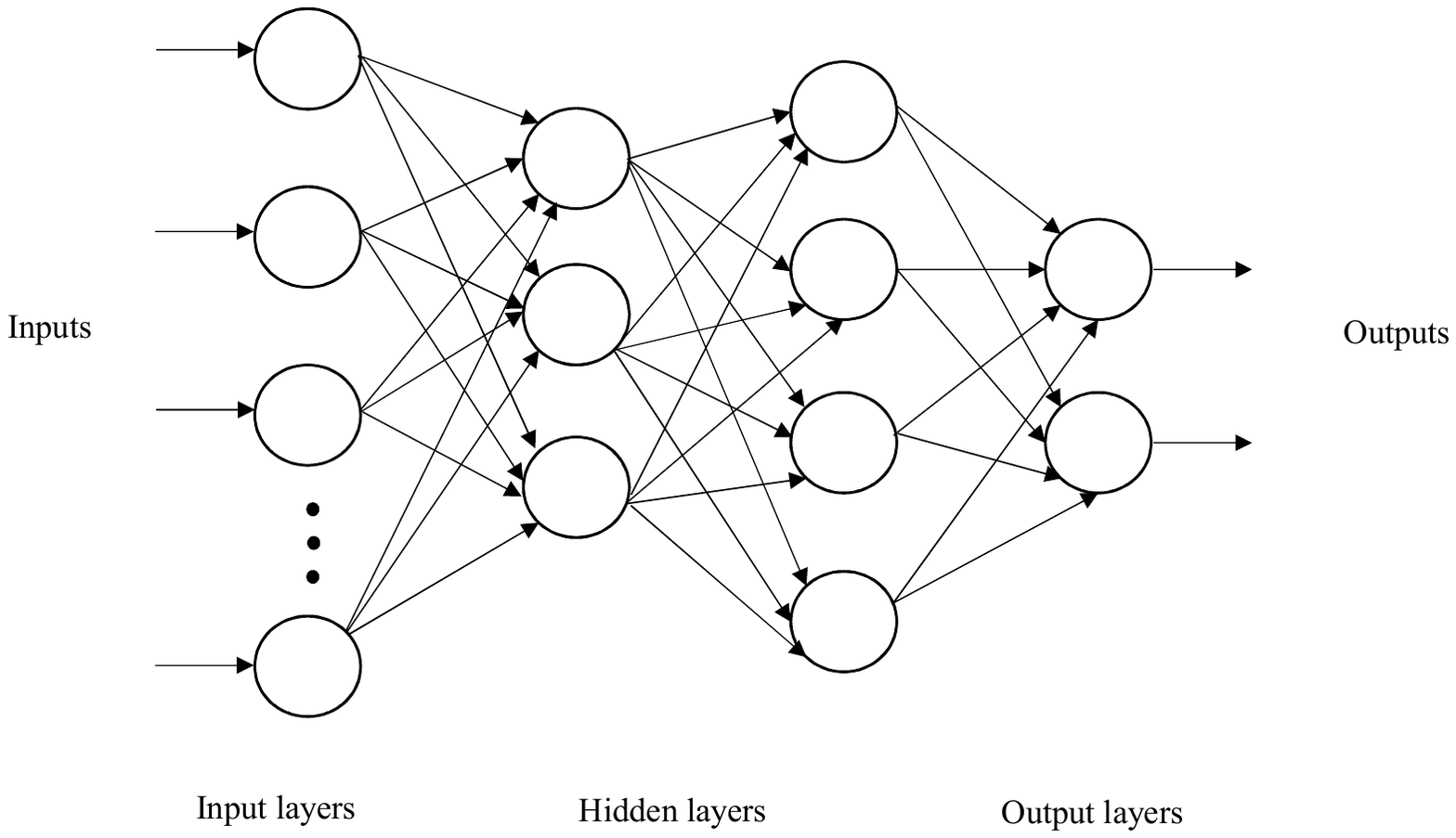}
\caption{\label{fig1}  A diagram for feed forward neural networks.} 
\end{figure}

A schematic diagram for feed forward neural networks is shown in Figure \ref{fig1}.

The architecture of neural networks are determined by the connection of nodes. This also determines how computation proceeds. In general terms, there are two type of neural networks: feed forward and feedback or recurrent networks. In the feed forward neural networks, the transfer of information only occurs in one direction, from the input nodes to the output nodes. There can be hidden nodes between input and output nodes. In feed forward neural networks, the output of any layer does not affect the same layer. In the recurrent neural networks, the transfer of information occurs in both direction by introducing loops in the network. The signals can go backward from output nodes to input nodes which makes recurrent neural networks complicated. 

The nodes of the graphs are called processing elements. These graphs are linked among themselves and these links are called connections. In principle, there is an infinite number of incoming connections that a processing element can receive and this is true for outgoing connections providing that signals in all of these are same. Processing elements can have local memory and they have transfer functions which can use local memory, can use input signals, and produce the processing element’s output signal. The transfer mechanism of signals is as follows: input signals to a neural network from outside the network arrive via connections that originate in the outside world \cite{30}.

\begin{figure}[H]
\centering
\includegraphics[width=5.8in]{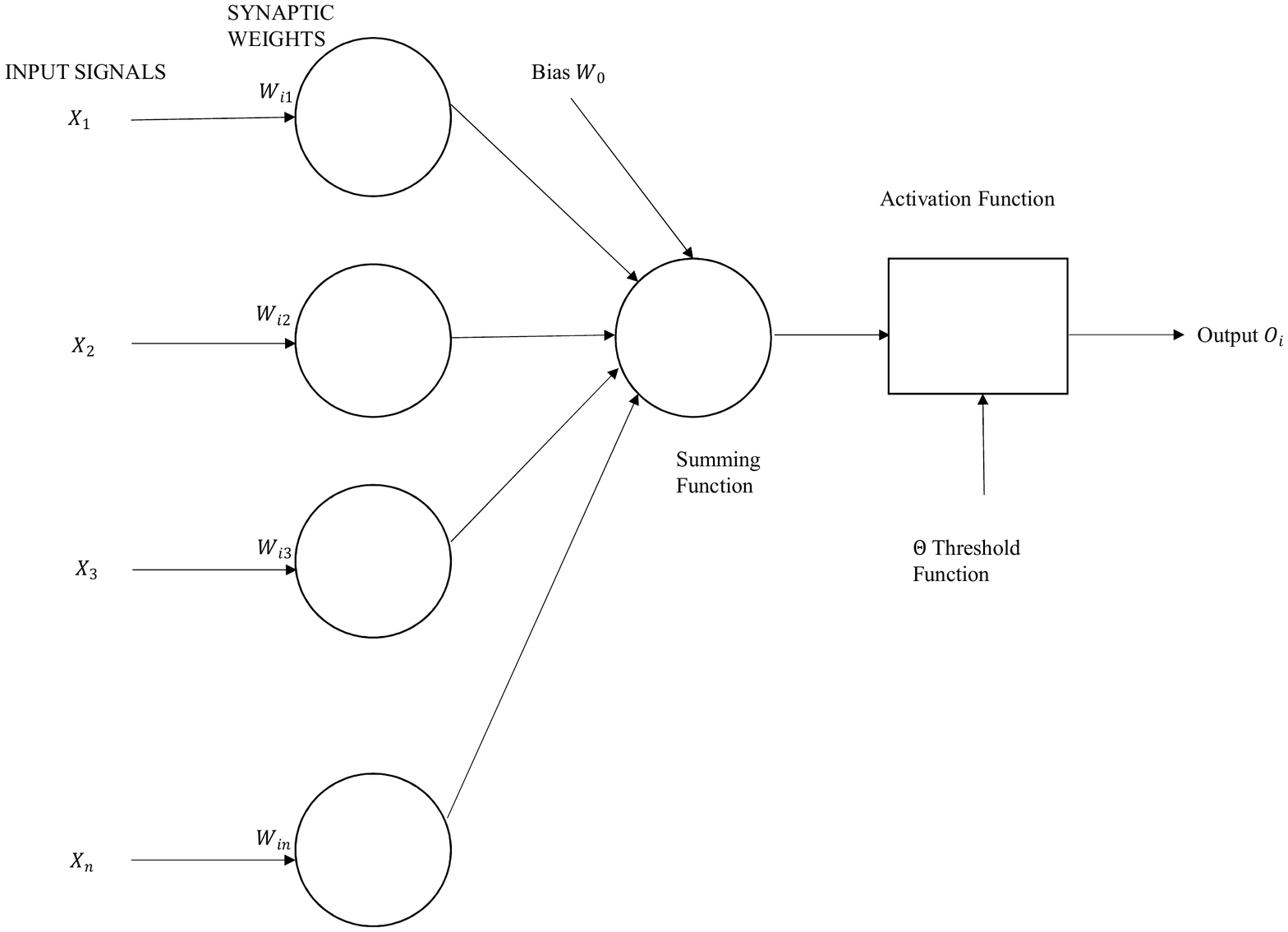}
\caption{\label{fig2} The mathematical model of artificial neural network.} 
\end{figure}

Input layers just send the information which comes from outside world to other layers. There is no process which contain usage of this information in input layers. Each neuron has just one input and one output. The hidden layers use the inputs (generally data) and produce outputs for the output layers. The output layers take the outputs from hidden layers, process data and produce output to the outside world. The mathematical model of artificial neural network is shown in Figure \ref{fig2}. 

A neuron (perceptron in computerized systems) $N_i$ accepts a set of $n$ inputs which is an element of the set $S=\left\lbrace x_j \vert j=1,2,\cdots,n \right\rbrace$. Each input is weighted before entrance of a neuron $N_i$ by weight factor $w_{ij}$ for $j=1,2,\cdots,n$. Furthermore, it has bias term $w_0$, a threshold value $\Theta_k$ which has to be reached or exceeded for the neuron to produce output signal. Mathematically, the output of the $i$-th neuron $N_i$ is
\begin{equation}
O_i=f (w_0+ \sum_{j=1}^n w_{ij}x_j).
\end{equation}
The neuron's $work$ condition can be defined as
\begin{equation}
w_0+ \sum_{j=1}^n w_{ij}x_j \geq \Theta,
\end{equation}
and the input signal of the $i$-th neuron $N_i$ is
\begin{equation}
s_i=w_0+\sum_{j=1}^n w_{ij}x_j.
\end{equation}
A function $f(s)$ acts on the produced weighted signal. This function is called activation function. The output signal obtained by activation function is 
\begin{equation}
O_i=f(s_i-\Theta_i).
\end{equation}
All inputs are multiplied by their weights and added together to form the net input to the neuron. This is called {\it net} and can be written as
\begin{equation}
\text{net}=\sum_{j=1} w_{ij}x_j+\Theta,
\end{equation}
where $\Theta$ is a threshold value which is added to the neurons. The neuron takes these inputs and produce an output $y$ by mapping function $f(net)$
\begin{equation}
y=f(\text{net})=f \left(  \sum_{j=1}^n  w_{ij}x_j+\Theta \right)
\end{equation}
where $f$ is the neuron activation function. Generally the neuron output function is proposed as a threshold function but linear, sign, sigmoid or step functions are widely used.

ANNs present new methods to approximately solve differential equations. The fundamental point which ANNs have is that, they are universal functions approximators. In differential equation theory, it is possible to use a trial function which is believed to solve the given differential equation. The choice of trial function can be so intuitive, maybe advanced guess. In the present paper, we used the method of Lagaris et al. \cite{18} in which a trial solution is proposed to ensure that initial or boundary conditions are satisfied. 

\subsection{Method of Trial Solution}
When the trial solution is employed, the ANNs output is replaced in the trial solution. The main role of ANNs is to generate a trial solution which approximates solutions of differential equations. In the work of Lagaris et al. \cite{18}, multilayer perceptron (MLP) neural network has been presented for the solution of both ordinary and partial differential equations. This method is based on function approximation capabilities of feed forward neural networks and yields in the construction of a solution. The difference between the constructed solution and exact solution results an error function. This concept uses a feed forward neural network as the basic approximation element, whose parameters (weights and biases) are adjusted to minimize an appropriate error function. To minimize error and to train network, optimization techniques are used which require error gradient computation with respect to inputs and neural network parameters. 

A differential equation defined on certain boundary conditions can be written as 
\begin{equation}
F \left( \vec{x}, y(\vec{x}), \nabla y(\vec{x}), \nabla^2 y(\vec{x}) \right)=0, ~ \vec{x} \in D, \label{dif1}
\end{equation}
where $\vec{x}=\left(x_1,x_2,\cdots,x_n \right) \in \mathbb{R}^n, ~ D \subset \mathbb{R}^n$ is the definition domain and $y(\vec{x})$ is the solution to be found. The solution of Eqn.(\ref{dif1}) is twofold consideration. On one side, one has to discretize the given domain $D$ and its boundary region $S$ into a set of points denoted as $\hat{D}$ and $\hat{S}$, respectively. After this discretization, Eqn.(\ref{dif1}) is transformed into
\begin{equation}
F \left( \vec{x}, y(\vec{x}), \nabla y(\vec{x}), \nabla^2 y(\vec{x}) \right)=0,~ \forall ~ \vec{x} \in \hat{D},
\end{equation}
under some specific boundary conditions. If $y_t(\vec{x},\vec{p})$ denotes a trial solution with the adjustable parameters $\vec{p}$, then the problem is transformed into an optimization problem 
\begin{equation}
\underset{\vec{p}}{\min} \sum_{x_i \in \vec{D}} F \left( \vec{x}_i, y_t(\vec{x}_i,\vec{p}), \nabla y_t(\vec{x}_i,\vec{p}), \nabla^2 y_t(\vec{x}_i,\vec{p}) \right)^2 \label{dif2}
\end{equation}
subject to some specific boundary conditions. In this consideration, the explicit or implicit form of the trial solution is not specified and trial solution $y_t$ employs a feed forward neural network and the parameters $\vec{p}$ correspond to the weights and biases of the neural architecture. On the other side of the consideration, one construct the trial function that can solve the differential equation. The general form of the trial solution can be written as the sum of two terms:
\begin{equation}
y_t(\vec{x})=A(\vec{x})+F(\vec{x},N(\vec{x},\vec{p})) \label{dif3}
\end{equation}
where $A(\vec{x})$ contains no adjustable parameters which satisfies the initial/boundary conditions and $N(\vec{x},\vec{p})$ is a single-output feed forward neural network with parameters $\vec{p}$ and $n$ input units fed with the input vector $\vec{x}$. $F$ is constructed in a way that it does not contribute to the boundary conditions since $y_t(\vec{x})$ must also satisfy them. This term is used to vanish the second term in Eqn.(\ref{dif3}) at the points of the initial or boundary conditions. $F$ employs a neural network whose weights and biases are to be adjusted in order to deal with the minimization problem. The problem has been reduced from the original constrained optimization problem to an unconstrained one due to the choice of the form of the trial solution that satisfies the boundary conditions \cite{18}. When the second term in Eqn.(\ref{dif3}) returns a zero value at initial or boundary conditions, trial solution satisfy automaticly these conditions. 

Solving a differential equation within ANN framework requires training of ANN. This training process consists of solving of the differential equation at any point in a given interval in which a solution is generally sought. Some of these points could not be considered during the training process. The minimization problem of Eqn.(\ref{dif2}) can be treated as a procedure of training the neural
network where the error corresponding to each input vector $\vec{x}_i$, $F(\vec{x}_i)$ has to become zero. Computation of this error value requires network output and the derivatives of the output with respect to any of its inputs. So the computing process of the gradient of the error with respect to the network weights involves computing not only the gradient of the network but also the gradient of the network derivatives with respect to its inputs \cite{18}.

To compute the gradient with respect to input vectors, consider a multilayer perceptron neural network with $n$ input units, one hidden layer with $H$ sigmoid units and a linear output unit. For a given input vector $\vec{x}=\left(x_1,x_2,\cdots,x_n \right)$, the output of the neural network is 
\begin{equation}
N(\vec{x},\vec{p})=\sum_{i=1}^H v_i \sigma(z_i) \label{nsum},
\end{equation}
where 
\begin{equation}
z_i=\sum_{j=1}^n w_{ij}x_j+u_i. \label{zi}
\end{equation}
In Eqn.(\ref{nsum}), $\sigma(z)$ is the sigmoid activation function, $v_i$ is weight from the hidden unit $i$ to the output  and in Eqn.(\ref{zi}), $w_{ij}$ denotes the weight from the input unit $j$ to hidden unit $i$ and $u_i$ is the bias of the hidden unit. In the present work, we used 
\begin{equation}
\sigma=\frac{1}{1+e^{-x}}
\end{equation}
as the activation function since it is possible to derive all the derivatives of $\sigma(x)$ in terms of itself. The activation function is called sigmoid function which is generally used in non-linear problems. It is also a monotonic function, i.e., either entirely non-increasing or non-decreasing. It makes easy for the model to generalize or adapt with variety of data and to differentiate between the output. This could be helpful when coding the related algorithm. $N$ is optimized to get an approximate solution and satisfy the given differential equation in the domain, not exactly but at least exactly. The ANNs output $N$ is then substituted in the trial solution. The related error function $E(x)$ can be defined as
\begin{equation}
E(x)=\sum_i \left\lbrace \frac{dy_t}{dx}-F(x,y)    \right\rbrace^2
\end{equation}
and to be minimized over all adjustable neural network parameters. In order to compute error function, derivatives of $N(\vec{x},\vec{p})$ with respect to $\vec{x}$ are needed:
\begin{equation}
\frac{\partial N}{\partial x_j}= \frac{\partial}{\partial x_j} \left[  \sum_i^H \nu_i \sigma \left( \sum_i^n w_{ij} x_j +u_i\right) \right] =\sum_i^h v_i w_{ij} \sigma^{(1)},
\end{equation}
where
\begin{equation}
\sigma^{(1)}=\frac{\partial \sigma(x)}{\partial x}.
\end{equation}
It is also possible to take $k$-th derivative of $N$:
\begin{equation}
\frac{\partial^k N}{\partial x_j^k}= \sum v_i w_{ij}^k \sigma_i^{(k)},
\end{equation}
where $\sigma_i=\sigma(z_i)$ and $\sigma_i^{(k)}$ denotes the $k$-th order derivative of the activation function. 

In order to minimize $E(x)$ or to make it zero if possible, it is necessary to find correct $x$ values. This can be done by training the neural network. Without knowing any solution point in advance, the neural network must be trained in an unsupervised procedure. The well-known back-propagation method is useless here since the error at each output unit is not available to the learning system. This makes target solution unavailable. A different optimization technique must be used, for example gradient descent method, in which the weights are initialized randomly. Therefore parameters of training need updating in the transfer process and this can be done with taking derivative of $N$ with respect to neural network parameters. The gradients of the single $N$ with respect to parameters are given as follows:
\begin{eqnarray}
\frac{dN}{dv_i}&=& \sigma(z_i) \\
\frac{dN}{du_i}&=& v_i \sigma(z_i)\\
\frac{dN}{dw_{ij}}&=&v_i \sigma(z_i)  x_j. 
\end{eqnarray} 
After computation of derivative of the error function with respect to the network parameters, these parameters are needed to be updated. The rule for this updating can be given as
\begin{eqnarray}
v_i(t+1)&=&v_i(t)+ \alpha \frac{\partial N_k}{\partial v_i} \\
u_i(t+1)&=& u_i(t)+ \beta \frac{\partial N_k}{\partial u_i} \\
w_{ij}(t+1) &=& w_{ij}(t) \gamma  \frac{\partial N_k}{\partial w_{ij}},
\end{eqnarray}
where $\alpha$, $\beta$ and $\gamma$ are the learning rates, $i=1,2,3,\cdots,n$ and $j=1,2,\cdots,h$. Once the derivative of the residual error function with respect to the neural network parameters has been defined, it is then straightforward to employ almost any minimization technique \cite{18}.

In view of the present discussion let us consider a second order differential equation
\begin{equation}
\frac{d^2y}{dx^2}=f \left( x,y,\frac{dy}{dx}  \right), ~y(a)=B, ~y(b)=C, ~x \in [a,b].
\end{equation}
We can write Eqn.(\ref{dif3}) as 
\begin{equation}
y_t(x,p)=A(x)+F(x)N(x,p)
\end{equation}
and define a trial function as 
\begin{equation}
y_t(x,p)=\tilde{a} x+\tilde{b}+(x-a)(x-b)N(x,p),
\end{equation}
where $\tilde{a}$ and $\tilde{b}$ are arbitrary parameters. The unique form of a trial solution is not specified but some conditions are need to be satisfied:
\begin{eqnarray}
A(a)=B, ~ A(b)=C \\
F(a)N(a,p)=0 \\
F(b)N(b,p)=0.\\
\end{eqnarray}
The error function for Blasius equation can be written as
\begin{equation}
E(x)=\sum_i^m \left\lbrace  \frac{d^3y_t(x,p)}{dx^3}+\frac{1}{2} y_t(x,p) \frac{d^2y_t(x,p)}{dx^2}   \right\rbrace^2
\end{equation}
where
\begin{equation}
y_t(x,p)=x^3+x^2+x^2(x-6)^2N(x,p)
\end{equation}
is the trial function for Blasius equation. There is not a unique function for determining the number of hidden layer and number of neurons in the hidden layers. It is tempting to create a simple network architecture for solving problem. Also less number of units could be chosen since trial and error beginning for one layer with few neurons present good network capabilities. In practice, using fewer hidden units can make generalization better. Furthermore, the training process should be easier because of fewer parameters are being used \cite{31}. In this present work, we have constructed a neural network with one input layer with one unit, one hidden layer with five units, and one output layer with one unit. The network was trained using a grid of varying equidistant points in given interval [0,a] where a changes according to the reference studies. To determine accuracy, training set is considered from inputs with 10  training points. The accuracy has been performed by conducting hundred independent runs and simulations results are chosen in average since training points are different in each  reference study which are compared.

\section{\label{sec:level3}Numerical Results}
\begin{figure}[H]
\includegraphics{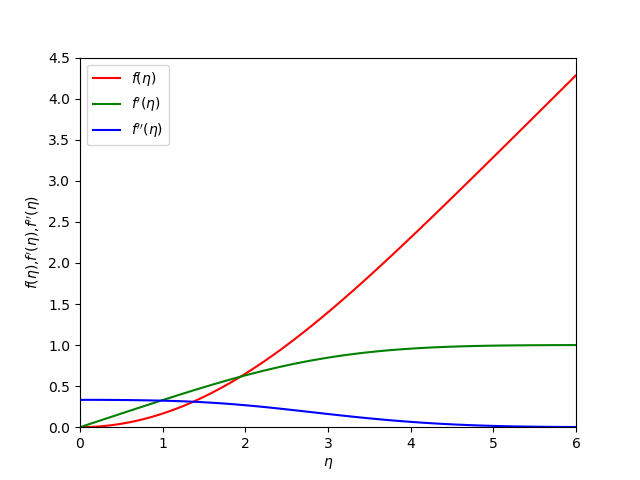}
\caption{\label{fig3} Plot of the  $f(\eta)$, $f^\prime(\eta)$ and $f^{\prime \prime}(\eta)$ for the Blasius flow.}
\end{figure}

Motivated with the previous discussion, we solved Blasius differential equation via feed forward neural networks. The graph of $f(\eta)$ and first and second derivatives are shown in Figure \ref{fig3}. We present our results in Tables \ref{tab:table1}, \ref{tab:table2}, \ref{tab:table3}, \ref{tab:table4}, \ref{tab:table5}, \ref{tab:table6}, \ref{tab:table7} and \ref{tab:table8}.  

We see from Fig.(\ref{fig3}) that the slope of $f^{\prime}(\eta)$ is being constant at large $\eta$ values. If the slope of a function is constant at large values, then the values of higher derivatives  must tend to zero which is the case for $f^{\prime \prime}(\eta)$ as can bee seen in Fig.(\ref{fig3}).

In Ref. \cite{12}, the authors proposed the sinc-collocation method for solving Blasius equation and obtained high accuracy solutions. They compared their results with the numerical results of \cite{3}. In Table \ref{tab:table1}, we compare our results for $f(\eta)$ with results of Ref. \cite{12} and give relative errors. 

\begin{table}[H]
\caption{\label{tab:table1} Comparison of $f(\eta)$ with \cite{3} and \cite{12}.}
\begin{ruledtabular}
\begin{tabular}{cccccccc}
$\eta$& $f(\eta)$ & \cite{3}& \cite{12} & Relative error with \cite{3} & Relative error with \cite{12}\\
\hline
0.2& 0.0066712  & 0.00664 & 0.0066429 & $4.70 \times 10^{-3}$ & $4.26 \times 10^{-3}$  \\
0.4& 0.0266161  & 0.02656 & 0.0266262 & $2.11 \times 10^{-3}$ & $3.79 \times 10^{-4}$  \\
0.6& 0.0598054  & 0.05973 & 0.0599956 & $1.26 \times 10^{-3}$ & $3.17 \times 10^{-3}$  \\
0.8& 0.1061811  & 0.10611 & 0.1057469 & $6.70 \times 10^{-4}$ & $4.10 \times 10^{-3}$ \\
1.0& 0.1656533  & 0.16557 & 0.1660500 & $5.03 \times 10^{-4}$ & $2.38 \times 10^{-3}$  \\ 
2.0& 0.6509571  & 0.65002 & 0.6499608 & $1.44 \times 10^{-3}$ & $1.53 \times 10^{-3}$  \\
3.0& 1.3981885  & 1.39681 & 1.3969174 & $9.86 \times 10^{-4}$ & $9.09 \times 10^{-4}$ \\
4.0& 2.3053710  & 2.30575 & 2.3058206 & $1.64 \times 10^{-4}$ & $1.94 \times 10^{-4}$ \\
5.0& 3.2827015  & 3.28327 & 3.2833274 & $1.73 \times 10^{-4}$ & $1.90 \times 10^{-4}$ \\
6.0& 4.2804686  & 4.27962 & 4.2796818 & $1.98 \times 10^{-4}$ & $1.83 \times 10^{-4}$ \\ 
7.0& 5.2794380  & 5.27924 & 5.2793263 & $3.75 \times 10^{-5}$ & $2.11 \times 10^{-5}$  \\
8.0& 6.2747581  & 6.27921 & 6.2792567 & $7.08 \times 10^{-4}$ & $7.16 \times 10^{-4}$ 
\end{tabular}
\end{ruledtabular}
\end{table}

The authors of Ref. \cite{17} solved Blasius equation by using predictor corrector two-point block method. The main motivation of their study was to provide a new method that can solve the higher order boundary value problem directly without reducing it to a system of first order equation. In Tables \ref{tab:table2}, \ref{tab:table3} and \ref{tab:table4} we compare our results of $f(\eta)$, $f^{\prime}(\eta)$ and $f^{\prime \prime}(\eta)$ respectively, with Ref. \cite{17} and reference therein. 

\begin{table}[H]
\caption{\label{tab:table2} Comparison of $f(\eta)$ with \cite{16} and \cite{17}.}
\begin{ruledtabular}
\begin{tabular}{ccccccccccc}
$\eta$& $f(\eta)$ & \cite{16}  & \cite{17} & Relative error with \cite{16} & Relative error with \cite{17} \\
\hline
0.0& 0.0000000  & 0.00000 & 0.000000000000 & - & - \\  
0.2& 0.0066712  & 0.00664 & 0.006640985327 & $4.69 \times 10^{-3}$ & $4.54 \times 10^{-3}$  \\
0.4& 0.0266161  & 0.02656 & 0.026559825361 & $2.11 \times 10^{-3}$ & $2.11 \times 10^{-3}$  \\
0.6& 0.0598054  & 0.05974 & 0.059734504720 & $1.09 \times 10^{-3}$ & $1.18 \times 10^{-3}$  \\
0.8& 0.1061811  & 0.10611 & 0.106107984345 & $6.70 \times 10^{-4}$ & $6.89 \times 10^{-4}$ \\
1.0& 0.1656533  & 0.16557 & 0.165571356468 & $5.03 \times 10^{-4}$ & $4.94 \times 10^{-4}$  \\ 
1.2& 0.2383573  & 0.23795 & 0.237948186814 & $1.71 \times 10^{-3}$ & $1.72 \times 10^{-3}$  \\
1.4& 0.3234848  & 0.32298 & 0.322980855006 & $1.56 \times 10^{-3}$ & $1.56 \times 10^{-3}$ \\
1.6& 0.4209526  & 0.42032 & 0.420319832654 & $1.50 \times 10^{-3}$ & $1.50 \times 10^{-3}$ \\
1.8& 0.5303212  & 0.52952 & 0.529516867107 & $1.51 \times 10^{-3}$ & $1.51 \times 10^{-3}$ \\
2.0& 0.6509571  & 0.65002 & 0.650022940030 & $1.44 \times 10^{-3}$ & $1.43 \times 10^{-3}$ \\ 
3.0& 1.3981885  & 1.39681 & 1.396805279908 & $9.86 \times 10^{-4}$ & $9.90 \times 10^{-4}$  \\
4.0& 2.3053710  & 2.30575 & 2.305741819331 & $1.64 \times 10^{-4}$ & $1.60 \times 10^{-4}$ \\  
5.0& 3.2827015  & 3.28327 & 3.283267477016 & $1.73 \times 10^{-4}$ & $1.72 \times 10^{-4}$ \\  
6.0& 4.2804686  & 4.27962 & 4.279613205851 & $1.98 \times 10^{-4}$ & $1.99 \times 10^{-4}$ \\  
7.0& 5.2794380  & 5.27924 & 5.279229586310 & $3.75 \times 10^{-5}$ & $3.94 \times 10^{-5}$ \\  
\end{tabular}
\end{ruledtabular}
\end{table}

\begin{table}[H]
\caption{\label{tab:table3} Comparison of $f^{\prime}(\eta)$ with \cite{16} and \cite{17}.}
\begin{ruledtabular}
\begin{tabular}{ccccccccccc}
$\eta$&$f^{\prime}(\eta)$ & \cite{16}  & \cite{17} & Relative error with \cite{16} & Relative error with \cite{17} \\
\hline
0.0& 0.0000000  & 0.00000 & 0.000000000000 & - & - \\  
0.2& 0.0665382  & 0.06641 & 0.066407645581 & $1.93 \times 10^{-3}$ & $1.96 \times 10^{-3}$  \\
0.4& 0.1328756  & 0.13276 & 0.132763864650 & $8.70 \times 10^{-4}$ & $8.41 \times 10^{-4}$  \\
0.6& 0.1989704  & 0.19894 & 0.198936807533 & $1.52 \times 10^{-4}$ & $1.68 \times 10^{-4}$  \\
0.8& 0.2647138  & 0.26471 & 0.264708546847 & $1.43 \times 10^{-5}$ & $1.98 \times 10^{-5}$ \\
1.0& 0.3298884  & 0.32978 & 0.329779295507 & $3.28 \times 10^{-4}$ & $3.30 \times 10^{-4}$  \\ 
1.2& 0.3941376  & 0.39378 & 0.393775229578 & $9.08 \times 10^{-4}$ & $9.20 \times 10^{-4}$  \\
1.4& 0.4569763  & 0.45627 & 0.456260757344 & $1.54 \times 10^{-3}$ & $1.56 \times 10^{-3}$ \\
1.6& 0.5175034  & 0.51676 & 0.516755653134 & $1.43 \times 10^{-3}$ & $1.44 \times 10^{-3}$ \\
1.8& 0.5757228  & 0.57476 & 0.574756899195 & $1.67 \times 10^{-3}$ & $1.68 \times 10^{-3}$ \\
2.0& 0.6309134  & 0.62977 & 0.629764390693 & $1.81 \times 10^{-3}$ & $1.82 \times 10^{-3}$ \\ 
3.0& 0.8450255  & 0.84604 & 0.846042808633 & $1.19 \times 10^{-4}$ & $1.20 \times 10^{-4}$  \\
4.0& 0.9564613  & 0.95552 & 0.955516599862 & $9.85 \times 10^{-4}$ & $9.88 \times 10^{-4}$ \\  
5.0& 0.9926906  & 0.99154 & 0.991540349003 & $1.16 \times 10^{-4}$ & $1.16 \times 10^{-4}$ \\  
6.0& 0.9973541  & 0.99897 & 0.998971359031 & $1.61 \times 10^{-3}$ & $1.61 \times 10^{-3}$ \\  
7.0& 0.9999295  & 0.99992 & 0.999920098970 & $9.50 \times 10^{-6}$ & $9.40 \times 10^{-6}$ \\  
\end{tabular}
\end{ruledtabular}
\end{table}

\begin{table}[H]
\caption{\label{tab:table4} Comparison of $f^{\prime \prime}(\eta)$ with \cite{16} and \cite{17}.}
\begin{ruledtabular}
\begin{tabular}{ccccccccccc}
$\eta$& $f^{\prime \prime}(\eta)$ & \cite{16}  & \cite{17} & Relative error with \cite{16} & Relative error with \cite{17} \\
\hline
0.0& 0.3327300  & 0.33206 & 0.332056697280 & $2.01 \times 10^{-3}$ & $2.02 \times 10^{-3}$ \\  
0.2& 0.3321784  & 0.33198 & 0.331983088239 & $5.97 \times 10^{-4}$ & $5.88 \times 10^{-4}$  \\
0.4& 0.3319617  & 0.33147 & 0.331469097613 & $1.48 \times 10^{-3}$ & $1.49 \times 10^{-3}$  \\
0.6& 0.3300398  & 0.33008 & 0.330078387197 & $1.21 \times 10^{-4}$ & $1.16 \times 10^{-4}$  \\
0.8& 0.3271891  & 0.32739 & 0.327388541771 & $6.13 \times 10^{-4}$ & $6.09 \times 10^{-4}$ \\
1.0& 0.3230290  & 0.32301 & 0.323006407767 & $5.88 \times 10^{-5}$ & $6.99 \times 10^{-5}$  \\ 
1.2& 0.3170408  & 0.31659 & 0.316588510401 & $1.42 \times 10^{-3}$ & $1.43 \times 10^{-3}$  \\
1.4& 0.3087311  & 0.30787 & 0.307864749053 & $2.80 \times 10^{-3}$ & $2.81 \times 10^{-3}$ \\
1.6& 0.2977530  & 0.29666 & 0.296662866548 & $3.68 \times 10^{-3}$ & $3.67 \times 10^{-3}$ \\
1.8& 0.2821543  & 0.28293 & 0.282930479596 & $2.74 \times 10^{-3}$ & $2.74 \times 10^{-3}$ \\
2.0& 0.2675027  & 0.26675 & 0.266751073418 & $2.82 \times 10^{-3}$ & $2.81 \times 10^{-3}$ \\ 
3.0& 0.1617696  & 0.16136 & 0.161360208525 & $2.54 \times 10^{-3}$ & $2.54 \times 10^{-3}$  \\
4.0& 0.0647938  & 0.06423 & 0.064234198123 & $8.77 \times 10^{-3}$ & $8.71 \times 10^{-3}$ \\  
5.0& 0.0158742  & 0.01591 & 0.015906860846 & $2.25 \times 10^{-3}$ & $2.05 \times 10^{-3}$ \\  
6.0& 0.0024080  & 0.00240 & 0.002402057604 & $3.33 \times 10^{-3}$ & $2.47 \times 10^{-3}$ \\  
7.0& 0.0002210  & 0.00022 & 0.000220171488 & $4.54 \times 10^{-3}$ & $3.76 \times 10^{-3}$ \\  
\end{tabular}
\end{ruledtabular}
\end{table}

In \cite{11}, the authors derive a short analytical expression by using the [4/3] Pade approximant for the derivative of the solution of Blasius equation. They obtained analytical and numerical solution for Blasius equation. In Table \ref{tab:table5}, we compare our results with \cite{11}.

\begin{table}[H]
\caption{\label{tab:table5} Comparison of $f(\eta)$ with \cite{11}.}
\begin{ruledtabular}
\begin{tabular}{ccccccccccc}
$\eta$& $f$ & \cite{11} Approximate  & \cite{11} Numerical & Relative error with \cite{11} (App.) & Relative error with \cite{11} (Num.) \\
\hline
0.0& 0.0000000  & 0.0    & 0.0    & - & - \\  
0.4& 0.0266161  & 0.0266 & 0.0266 & $6.05 \times 10^{-4}$ & $6.05 \times 10^{-4}$  \\
0.8& 0.1061811  & 0.1061 & 0.1061 & $7.64 \times 10^{-4}$ & $1.49 \times 10^{-4}$  \\
1.2& 0.2383573  & 0.2379 & 0.2379 & $1.92 \times 10^{-3}$ & $1.92 \times 10^{-3}$  \\
1.6& 0.4209526  & 0.4203 & 0.4203 & $1.55 \times 10^{-3}$ & $1.55 \times 10^{-3}$ \\
2.0& 0.6509571  & 0.6500 & 0.6500 & $1.47 \times 10^{-3}$ & $1.47 \times 10^{-3}$  \\ 
2.4& 0.9237037  & 0.9223 & 0.9223 & $1.52 \times 10^{-3}$ & $1.52 \times 10^{-3}$  \\
2.8& 1.2324979  & 1.2311 & 1.2310 & $1.13 \times 10^{-3}$ & $1.21 \times 10^{-3}$ \\
3.2& 1.5714499  & 1.5693 & 1.5691 & $1.36 \times 10^{-3}$ & $1.49 \times 10^{-3}$ \\
3.6& 1.9298989  & 1.9297 & 1.9295 & $1.03 \times 10^{-4}$ & $2.06 \times 10^{-4}$ \\
4.0& 2.3053710  & 2.3058 & 2.3057 & $1.86 \times 10^{-4}$ & $1.42 \times 10^{-4}$ \\ 
4.4& 2.6915585  & 2.6922 & 2.6924 & $2.38 \times 10^{-4}$ & $3.12 \times 10^{-4}$  \\
4.6& 2.8874084  & 2.8879 & 2.8882 & $1.70 \times 10^{-4}$ & $2.74 \times 10^{-4}$ \\  
4.8& 3.0845629  & 3.0848 & 3.0853 & $7.68 \times 10^{-5}$ & $2.38 \times 10^{-4}$ \\  
5.0& 3.2827015  & 3.2827 & 3.2833 & $4.56 \times 10^{-7}$ & $1.82 \times 10^{-4}$ \\  
5.2& 3.4863866  & 3.4813 & 3.4819 & $1.46 \times 10^{-3}$ & $1.28 \times 10^{-3}$ \\  
5.4& 3.6856982  & 3.6805 & 3.6809 & $1.42 \times 10^{-3}$ & $1.30 \times 10^{-3}$ \\  
5.6& 3.8853086  & 3.8799 & 3.8803 & $1.39 \times 10^{-3}$ & $1.29 \times 10^{-3}$ \\  
5.8& 4.0851186  & 4.0796 & 4.0799 & $1.35 \times 10^{-3}$ & $1.27 \times 10^{-3}$ \\  
6.0& 4.2804686  & 4.2794 & 4.2796 & $2.49 \times 10^{-4}$ & $2.02 \times 10^{-4}$ \\  
6.4& 4.6852511  & 4.6793 & 4.6794 & $1.27 \times 10^{-3}$ & $1.25 \times 10^{-3}$ \\  
6.8& 5.0858307  & 5.0792 & 5.0793 & $1.30 \times 10^{-3}$ & $1.28 \times 10^{-3}$ \\  
7.0& 5.2794380  & 5.2792 & 5.2792 & $4.50 \times 10^{-5}$ & $4.50 \times 10^{-5}$ \\  
7.4& 5.6878210  & 5.6792 & 5.6792 & $1.52 \times 10^{-3}$ & $1.52 \times 10^{-3}$ \\  
8.0& 6.2747581  & 6.2792 & 6.2792 & $7.07 \times 10^{-4}$ & $7.07 \times 10^{-4}$ \\  
10 & 8.3094360  & 8.2792 & 8.2792 & $3.65 \times 10^{-3}$ & $3.65 \times 10^{-3}$ \\  
20 & 18.046656  & 18.2792 & 18.2792 & $1.27 \times 10^{-2}$ & $1.27 \times 10^{-2}$ \\  
100& 95.536872  & 98.2792 & 98.2792 & $2.80 \times 10^{-2}$ & $2.80 \times 10^{-2}$ \\  

\end{tabular}
\end{ruledtabular}
\end{table}

In Ref. \cite{5}, the authors solved Blasius equation via differential transformation method. In Tables \ref{tab:table6}, \ref{tab:table7} and \ref{tab:table8}, we compare our results of $f(\eta)$, $f^{\prime}(\eta)$ and  $f^{\prime \prime}(\eta)$ respectively, with the results of Ref. \cite{5}.

\begin{table}[H]
\caption{\label{tab:table6} Comparison of $f(\eta)$ with \cite{5}.}
\begin{ruledtabular}
\begin{tabular}{ccccccccccc}
$\eta$& $f(\eta)$ & \cite{5}& Relative error with \cite{5}  \\
\hline
0.0& 0.0000000  & 0   & -   \\  
0.5& 0.0415583  & 0.0414928 & $1.57 \times 10^{-3}$  \\
1.0& 0.1656533  & 0.1655716 & $4.93 \times 10^{-4}$  \\
1.5& 0.3704523  & 0.3701382 & $8.49 \times 10^{-4}$  \\
2.0& 0.6509571  & 0.6500239 & $1.53 \times 10^{-3}$  \\
2.5& 0.9977967  & 0.9963104 & $1.49 \times 10^{-3}$   \\ 
3.0& 1.3981885  & 1.3968070 & $9.89 \times 10^{-4}$  \\
3.5& 1.8382739  & 1.8376970 & $3.14 \times 10^{-4}$  \\
4.0& 2.3053710  & 2.3057450 & $1.62 \times 10^{-4}$ \\
4.5& 2.7892980  & 2.7901320 & $2.99 \times 10^{-4}$  \\
5.0& 3.2827015  & 3.2832720 & $1.73 \times 10^{-4}$ \\ 
5.5& 3.7807347  & 3.7805700 & $4.36 \times 10^{-5}$   \\
6.0& 4.2804686  & 4.2796190 & $1.98 \times 10^{-4}$ \\  
6.5& 4.7802949  & 4.7793210 & $2.03 \times 10^{-4}$ \\  
7.0& 5.2794380  & 5.2792370 & $3.81 \times 10^{-5}$ \\  
7.5& 5.7776055  & 5.7792170 & $2.79 \times 10^{-4}$ \\  
8.0& 6.2747581  & 6.2792120 & $7.10 \times 10^{-4}$ \\  
8.5& 6.7709708  & 6.7792110 & $1.21 \times 10^{-3}$ \\  
9.0& 7.2663553  & 7.2792110 & $1.76 \times 10^{-3}$ \\  

\end{tabular}
\end{ruledtabular}
\end{table}

\begin{table}[H]
\caption{\label{tab:table7} Comparison of $f^{\prime}(\eta)$  with \cite{5}.}
\begin{ruledtabular}
\begin{tabular}{ccccccccccc}
$\eta$& $f^{\prime}(\eta)$ & \cite{5}& Relative error with \cite{5}  \\
\hline
0.0& 0.0000000  & 0   & -   \\  
0.5& 0.1659582  & 0.1658851 & $4.41 \times 10^{-4}$  \\
1.0& 0.3298884  & 0.3297798 & $3.29 \times 10^{-4}$  \\
1.5& 0.4876893  & 0.4867890 & $1.85 \times 10^{-3}$  \\
2.0& 0.6309134  & 0.6297654 & $1.83 \times 10^{-3}$  \\
2.5& 0.7518429  & 0.7512593 & $7.77 \times 10^{-4}$   \\ 
3.0& 0.8450255  & 0.8460440 & $1.20 \times 10^{-3}$  \\
3.5& 0.9110446  & 0.9130400 & $2.18 \times 10^{-3}$  \\
4.0& 0.9564613  & 0.9555179 & $9.87 \times 10^{-4}$ \\
4.5& 0.9793287  & 0.9795140 & $1.89 \times 10^{-4}$  \\
5.0& 0.9926906  & 0.9915417 & $1.16 \times 10^{-3}$ \\ 
5.5& 0.9984863  & 0.9968786 & $1.61 \times 10^{-3}$   \\
6.0& 0.9973541  & 0.9989727 & $1.62 \times 10^{-3}$ \\  
6.5& 0.9997284  & 0.9996988 & $2.96 \times 10^{-5}$ \\  
7.0& 0.9999295  & 0.9999214 & $8.10 \times 10^{-6}$ \\  
7.5& 0.9999426  & 0.9999816 & $3.90 \times 10^{-5}$ \\  
8.0& 0.9999647  & 0.9999959 & $3.12 \times 10^{-5}$ \\  
8.5& 0.9999779  & 0.9999989 & $2.1 \times 10^{-5}$ \\  
9.0& 0.9999991  & 0.9999995 & $4. \times 10^{-7}$ \\  

\end{tabular}
\end{ruledtabular}
\end{table}

\begin{table}[H]
\caption{\label{tab:table8} Comparison of $f^{\prime \prime}(\eta)$  with \cite{5}.}
\begin{ruledtabular}
\begin{tabular}{ccccccccccc}
$\eta$& $f^{\prime \prime}(\eta)$ & \cite{5}& Relative error with \cite{5}  \\
\hline
0.0& 0.3327300  & 0.3320571 & $2.03 \times 10^{-3}$    \\  
0.5& 0.3304948  & 0.3309107 & $1.25 \times 10^{-3}$  \\
1.0& 0.3230290  & 0.3230069 & $6.84 \times 10^{-5}$  \\
1.5& 0.3043979  & 0.3025803 & $6.00 \times 10^{-3}$  \\
2.0& 0.2675027  & 0.2667514 & $2.82 \times 10^{-3}$  \\
2.5& 0.2145581  & 0.2174115 & $1.31 \times 10^{-2}$   \\ 
3.0& 0.1617696  & 0.1613603 & $2.54 \times 10^{-3}$  \\
3.5& 0.1071664  & 0.1077726 & $5.62 \times 10^{-3}$  \\
4.0& 0.0647938  & 0.0642341 & $8.71 \times 10^{-3}$ \\
4.5& 0.0370600  & 0.0339809 & $9.06 \times 10^{-2}$  \\
5.0& 0.0158742  & 0.0159068 & $2.05 \times 10^{-3}$ \\ 
5.5& 0.0063681  & 0.0065786 & $3.20 \times 10^{-2}$   \\
6.0& 0.0024080  & 0.0024020 & $2.50 \times 10^{-3}$ \\  
6.5& 0.0007749  & 0.0007741 & $1.03 \times 10^{-3}$ \\  
7.0& 0.0002210  & 0.0002202 & $3.63 \times 10^{-3}$ \\  
7.5& 0.0000558  & 0.0000553 & $9.04 \times 10^{-3}$ \\  
8.0& 0.0000128  & 0.0000122 & $4.91 \times 10^{-2}$ \\  
8.5& 0.0000034  & 0.0000024 & $4.17 \times 10^{-1}$ \\  
9.0& 0.0000009  & 0.0000005 & $8. \times 10^{-1}$ \\  

\end{tabular}
\end{ruledtabular}
\end{table}

\section{\label{sec:level4}Discussion and Conclusion}
Various methods have been used to solve differential equations analytically and approximately. With the aid of computers, neural network methods based on multilayer perceptron, radial basis functions and finite element etc. are presented for solving differential equations as alternative methods.

Blasius equation has a great attention and various approaches have been presented to solve it. The main problem in this area is the accuracy and range of applicability of these approaches. In the present paper, we studied Blasius equation via feed forward neural network. A comprehensive comparison with previous studies is presented. This comparison shows that the absolute error of the proposed scheme lies in the range of $10^{-7}-10^{-1}$ with other studies which is an inducement of the validation of feed forward neural network as an alternative method to solve Blasius equation. In our approach, we did not convert Blasius equation into a system of first order differential equations which is general attitude for higher order differential equations.

\begin{acknowledgments}
This research did not receive any specific grant from funding agencies in the public, commercial,
or not-for-profit sectors.
\end{acknowledgments}

\end{document}